%% file: iclr2018_conference.tex
\documentclass{article} 
\usepackage{iclr2018_conference,times}
\usepackage{hyperref}
\usepackage{url}
\usepackage{graphicx}
\usepackage{amssymb,amsmath}
\usepackage{bm}

\input{math_commands.tex}

\title{Gradient Masking Causes CLEVER to Overestimate Adversarial Perturbation Size}

\author{Ian Goodfellow \\
Google Brain\\
Mountain View, CA\\
goodfellow@google.com}

\iclrfinalcopy 

\begin{document}

\maketitle

\begin{abstract}
A key problem in research on adversarial examples is that vulnerability
to adversarial examples is usually measured by running attack algorithms.
Because the attack algorithms are not optimal, the attack algorithms are
prone to overestimating the size of perturbation needed to fool the
target model.
In other words, the attack-based methodology provides an upper-bound on
the size of a perturbation that will fool the model, but security guarantees
require a lower bound.
CLEVER is a proposed scoring method to estimate a lower bound.
Unfortunately, an estimate of a bound is not a bound.
In this report, we show that gradient masking, a common problem that causes
attack methodologies to provide only a very loose upper bound, causes
CLEVER to overestimate the size of perturbation needed to fool the model.
In other words, CLEVER does not resolve the key problem with the attack-based
methodology, because it fails to provide a lower bound.
\end{abstract}

\section{Introduction}

A key problem in research on adversarial examples is that vulnerability
to adversarial examples is usually measured by running attack algorithms
\citep{Szegedy-ICLR2014}.
Because the attack algorithms are not optimal, the attack algorithms are
prone to overestimating the size of perturbation needed to fool the
target model.
In other words, the attack-based methodology provides an upper-bound on
the size of a perturbation that will fool the model, but security guarantees
require a lower bound.
CLEVER \citep{weng2018evaluating} is a proposed scoring method to estimate a lower bound.
In this report, we show that gradient masking, a common problem that causes
attack methodologies to provide only a very loose upper bound, causes
CLEVER to overestimate the size of perturbation needed to fool the model.
In other words, CLEVER does not resolve the key problem with the attack-based
methodology, because it fails to provide a lower bound.

\section{CLEVER}

CLEVER is based on estimating the local Lipschitz constant of a model represented
by a function $f$ applied to an input $\vx$.
The estimate is formed by observing $p$-norms of the gradient,
$\nabla_\vx \| f(\vx) \|_p$ at $N_s$ random sampled points
$\vx$. These observations are then used to form a statistical estimate of the local
Lipschitz constant.

CLEVER is only intended to be applied to Lipschitz-continous functions.

\section{Counterexample allowing for infinite precision}
\label{sec:theory}

First, we show that CLEVER can underestimate the local Lipschitz constant,
even in a theoretical setting, where we assume CLEVER is able to represent real
numbers with unlimited precision.

Suppose that we have some function $g$ which has a large local Lipschitz constant
at some input $\vx$. For example, $g$ could be a typical neural network with
no defenses against adversarial examples.

Define $f(\vx) = g(h(\vx)),$ where $h(\vx)$ is a staircase function, rounding the
input to a set of quantized values. One example of such a staircase function
is
\[ h(x) = \frac{\mathrm{ceil}( c x)} {c}, \]
where $c$ is a hyperparameter determining how many levels to quantize
each unit interval into.

Because the function $h$ has zero gradient ``almost everywhere''
our $N_s$ samples of $\vx$ will all observe zero gradient of $f$ ``almost surely''.
Here ``almost everywhere'' and ``almost surely'' are used in the measure theoretic
sense, meaning that they hold except for a set of measure zero.

Having observed no gradient of $f$, the CLEVER score will regard $f$ as constant,
even though by construction (we assumed $g$ is highly sensitive to its input)
it is not. This shows that CLEVER will conclude the model is highly robust,
even though it is not.

\begin{figure}
  \includegraphics{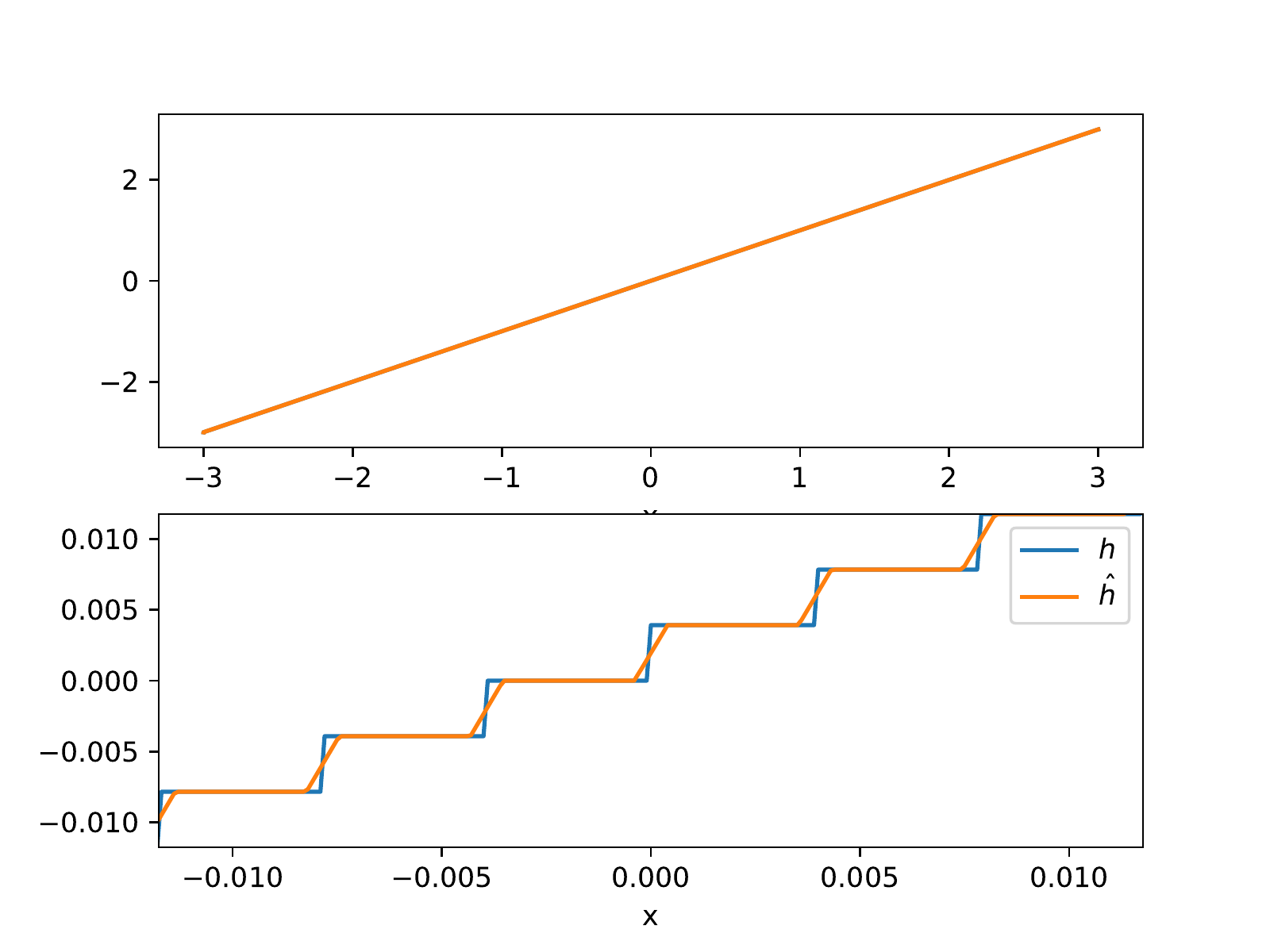}
  \caption{
    The staircase function $h$ with $c = 255$ steps per unit interval, and the Lipschitz continuous approximation $\hat{h}$ using linear ramps of width $\delta$ set to $5 \times$ smaller than the step width.
 In the upper, zoomed out view, both $h$ and $\hat{h}$ resemble the
  identity function.
  In the lower, zoomed in view, we see that $h$ in fact has a derivative
  of zero almost everywhere.
  The Lipschitz continuous approximation of $\hat{h}$ has a derivative of
  zero with probability $1 - \delta$ at points sampled uniformly at random.
  By shrinking $\delta$ further we can make the probability of observing
  nonzero derivatives arbitrarily small.
  Inserting this function anywhere in a model will not interfere with the
  operation of the model (beyond reducing its precision to $8$ bits for
  $c=255$) but will prevent CLEVER from observing nonzero gradient with
  arbitrarily high probability.
  }
  \label{fig:ramps}
\end{figure}

There is one problem with this example though: the function $f$ is not Lipschitz
continuous. CLEVER is intended only for Lipschitz continuous functions.
This difficulty can be resolved by approximating the staircase function $h$
with a Lipschitz continuous function $\hat{h}$.
For example, instead of a staircase function that increases with linear jumps,
use a staircase-like function that increases with linear ramps of width $\delta$,
as illustrated in Figure \ref{fig:ramps}.
As $\delta\rightarrow 0$, the probability of CLEVER observing a non-zero gradient
with finite $N_S$ becomes arbitrarily small.

\section{Implementation on Digital Computers}
\label{sec:practice}

The problems in the previous section are applicable even if CLEVER is able
to represent real numbers with unlimited precision.

In practice, because CLEVER is a metric to be used in experiments,
CLEVER must be regarded as a form of software that executes on a digital
computer to evaluate finite-precision machine learning models, rather
than as an abstract function applied to abstract
real-valued variables.

In a digital computer, every function is either constant or it is not
Lipschitz continuous, because the rounding to a finite number of bits
means that any function that increases or decrease does so in discrete
jumps. Because of this, it is not possible to satisfy the conditions
of the theory for CLEVER in actual usage.

When CLEVER is used on a digital computers, it can be prone to various
difficult-to-predict failures.
For example, sigmoid units in the network may saturate to the point
that the gradient through them is numerically rounded to machine
zero.
This would cause an effect similar to the staircase example above,
with the sharp sigmoid function appearing not to be Lipschitz continuous
due to numerical error, and the function appearing to CLEVER to
be extremely robust due to the lack of observed gradient.
Numerical saturation of sigmoid units has already been observed to
interfere with attack-based benchmarking of defenses against adversarial
examples \citep{brendel2017comment} and CLEVER does not resolve this problem.

\section{Silent failures}

CLEVER is not intended for use on functions that are not Lipschitz continuous,
but numerical error in digital computers can cause functions that are Lipschitz
continuous in theory to be far from Lipschitz continuous in practice.
CLEVER does not offer a mechanism to detect when this happens.
It may be possible to get reasonable estimates from CLEVER if the user of CLEVER
is aware of the problems causing loss of gradient and can mitigate them
(e.g., by removing the staircase function $h$ proposed in Section \ref{sec:theory}).
Unfortunately, inaccuracies resulting from numerical error are difficult to
characterize, anticipate, or detect, so a user of CLEVER who obtains a good
score will not know whether the model is robust or whether CLEVER has returned
an inaccurate estimate without raising a warning that the method is not
applicable to the current model.

\section{Conclusion}

The use of the staircase function described in Section \ref{sec:theory}
and the various numerical difficulties described in Section \ref{sec:practice}
are all examples of {\em gradient masking } \citep{papernot2017practical}.
Gradient masking is any defense against adversarial examples that works by
breaking attack algorithms by making the gradient useless (small or pointed
in the wrong direction, too noisy, combined with a poorly conditioned Hessian,
etc.).
A key flaw in the methodology of evaluating defenses against adversarial examples
by testing them against attacks is that the attacks can fail when the defender
(intentionally or unintentionally, knowingly or unknowingly) uses gradient masking.
This report shows that CLEVER suffers from the same flaw as the attack-based
methodology, and does not actually offer a lower bound on the size of perturbation
required to fool the model.

\bibliography{iclr2018_conference}
\bibliographystyle{iclr2018_conference}

\end{document}

%% file: math_commands.tex
\def\eqref#1{equation~\ref{#1}}

\def\1{\bm{1}}

\def\vx{{\bm{x}}}

\newif\ifMIT
\MITfalse
\ifMIT

\def\gamma{{\boldmath{\vgamma}}}

\else

\def\vgamma{{\bm{\gamma}}}

\fi

\ifMIT

\else

\fi

\DeclareMathAlphabet{\mathsfit}{\encodingdefault}{\sfdefault}{m}{sl}
\SetMathAlphabet{\mathsfit}{bold}{\encodingdefault}{\sfdefault}{bx}{n}

\ifMIT
\else

\fi

%% file: iclr2018_conference.bbl
\begin{thebibliography}{4}
\providecommand{\natexlab}[1]{#1}
\providecommand{\url}[1]{\texttt{#1}}
\expandafter\ifx\csname urlstyle\endcsname\relax
  \providecommand{\doi}[1]{doi: #1}\else
  \providecommand{\doi}{doi: \begingroup \urlstyle{rm}\Url}\fi

\bibitem[Brendel \& Bethge(2017)Brendel and Bethge]{brendel2017comment}
Wieland Brendel and Matthias Bethge.
\newblock Comment on ``biologically inspired protection of deep networks from
  adversarial attacks''.
\newblock \emph{arXiv preprint arXiv:1704.01547}, 2017.

\bibitem[Papernot et~al.(2017)Papernot, McDaniel, Goodfellow, Jha, Celik, and
  Swami]{papernot2017practical}
Nicolas Papernot, Patrick McDaniel, Ian Goodfellow, Somesh Jha, Z~Berkay Celik,
  and Ananthram Swami.
\newblock Practical black-box attacks against machine learning.
\newblock In \emph{Proceedings of the 2017 ACM on Asia Conference on Computer
  and Communications Security}, pp.\  506--519. ACM, 2017.

\bibitem[Szegedy et~al.(2014)Szegedy, Zaremba, Sutskever, Bruna, Erhan,
  Goodfellow, and Fergus]{Szegedy-ICLR2014}
Christian Szegedy, Wojciech Zaremba, Ilya Sutskever, Joan Bruna, Dumitru Erhan,
  Ian~J. Goodfellow, and Rob Fergus.
\newblock Intriguing properties of neural networks.
\newblock \emph{ICLR}, abs/1312.6199, 2014.
\newblock URL \url{http://arxiv.org/abs/1312.6199}.

\bibitem[Weng et~al.(2018)Weng, Zhang, Chen, Yi, Su, Gao, Hsieh, and
  Daniel]{weng2018evaluating}
Tsui-Wei Weng, Huan Zhang, Pin-Yu Chen, Jinfeng Yi, Dong Su, Yupeng Gao,
  Cho-Jui Hsieh, and Luca Daniel.
\newblock Evaluating the robustness of neural networks: An extreme value theory
  approach.
\newblock In \emph{International Conference on Learning Representations}, 2018.
\newblock URL \url{https://openreview.net/forum?id=BkUHlMZ0b}.

\end{thebibliography}
